# Lightweight Image Super-Resolution with Adaptive Weighted Learning Network

Chaofeng Wang Zheng Li Jun Shi
Shanghai University
Shanghai, China

## Abstract

*Deep learning has been successfully applied to the single-image super-resolution (SISR) task with great performance in recent years. However, most convolutional neural network based SR models require heavy computation, which limit their real-world applications. In this work, a lightweight SR network, named Adaptive Weighted Super-Resolution Network (AWSRN), is proposed for SISR to address this issue. A novel local fusion block (LFB) is designed in AWSRN for efficient residual learning, which consists of stacked adaptive weighted residual units (AWRU) and a local residual fusion unit (LRFU). Moreover, an adaptive weighted multi-scale (AWMS) module is proposed to make full use of features in reconstruction layer. AWMS consists of several different scale convolutions, and the redundancy scale branch can be removed according to the contribution of adaptive weights in AWMS for lightweight network. The experimental results on the commonly used datasets show that the proposed lightweight AWSRN achieves superior performance on ×2, ×3, ×4, and ×8 scale factors to state-of-the-art methods with similar parameters and computational overhead.Code is avaliable at: https://github.com/ChaofWang/AWSRN*

## 1. Introduction

Single image super-resolution (SISR) is a classic computer vision task which reconstructs a high resolution (HR) image from a low resolution (LR) one. Although a number of solutions have been proposed for SISR, it is still a challenging task as an ill-posed inverse procedure.

With the fast development of deep learning (DL) technology, it has been successfully applied to SR task. After Dong et al. proposed a convolutional neural networks (CNN) based SR (SRCNN) algorithm that outperformed its previous work [1], various improved algorithms have been proposed for SISR with superior performance [2,4,5].

It is well known that the deeper networks generally achieve better performance, especially in the residual learning (RL) [4]. Therefore, many very deep SR networks have been proposed with state-of-the-art performance, such as EDSR [6], RDN [7], RCAN [8]. However, these SR networks generally suffer from the problem of a heavy burden on computational resources with large model sizes [9], which limits their wide real-world applications. Consequently, the design of lightweight CNN for SISR has attracted considerable attention recently.

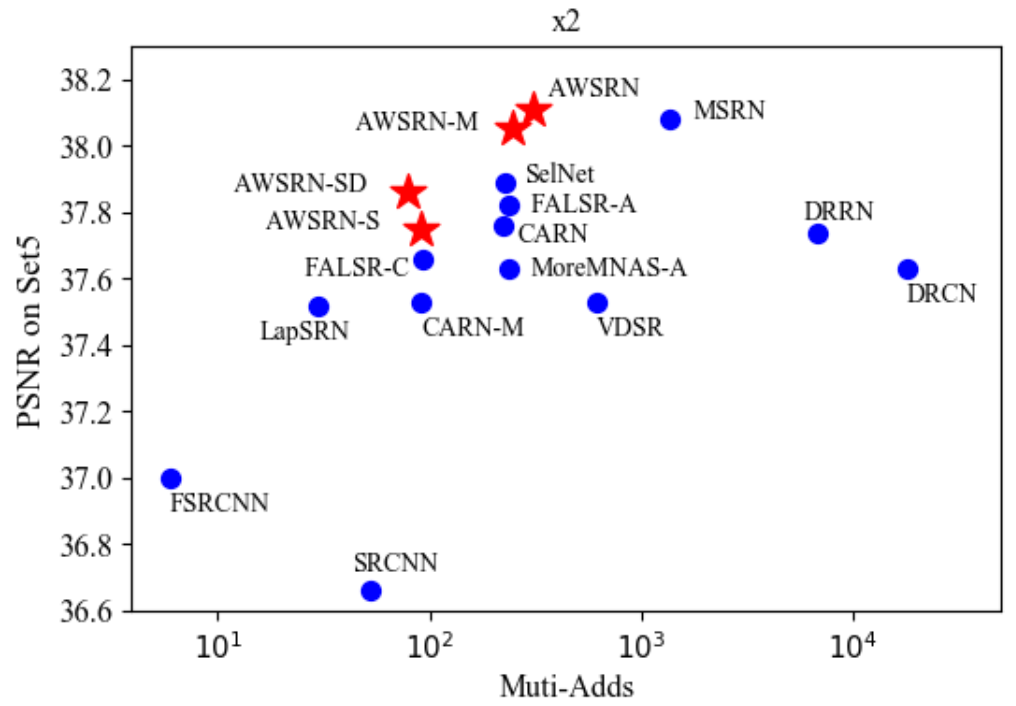

Figure 1. Performance comparison between our AWSRN family (red star) and other state-of-the-art lightweight networks (blue circle) on Set5 with a scale factor of 2. Note that the Muti-Adds represents the number of operations, and the output image size is 1280×720 to calculate Muti-Adds in this work.

The popular way to build a lightweight network is to reduce the number of model parameters. A simple strategy is to construct a shallow network model for SISR, such as ESPCN [10] and FSRCNN [3] algorithms. Another approach is to share parameters through recursive mechanisms. DRRN [11] and DRCN [5] are two typical networks of this method for SISR with less network parameters but better performance. However, the numbers of computational operations (Multi-Adds) of these networks are still very large [9]. In addition, neural architecture search (NAS) is an emerging approach to automatically design efficient networks [12], which is then introduced to the SR task to develop the MoreMNA-S and FALSR algorithms [13,14]. As shown in Figure 1, both MoreMNA-S and FALSR algorithms have fewer computational operations. However, due to the constraints of search space and strategy in NAS, the performance of NAS-based networks is also limited.

On the other hand, the very deep SR networks, especially the RL-based SR networks, easily suffer from the exploding gradient problem, resulting in instability for SR network training. The trick of residual scaling is then proposed to alleviate this issue [6]. Residual scaling [16] aims to select fixed weights value for residual units to limit the gradient flow. However, this trick cannot guarantee the improvement of performance for SR networks. Moreover, most SR networks only have a single-scale reconstruction layer with convolution, transposed convolution or subpixelshuffle [10], resulting in insufficiently used feature information from the nonlinear mapping module. Although the way of multi-scale reconstruction provides more information, it induces more parameters which result in large number of computational operations.

In this work, we propose a lightweight adaptive weighted SR network (AWSRN) for SISR as shown in Figure 2. AWSRN consists of a feature extraction module, a nonlinear mapping module and an adaptive weight multi-scale (AWMS) reconstruction module. Our AWSRN achieves state-of-the-art performance for SR but with fewer network parameters as shown in Figure 1.

The main contributions of this work are threefold:

(1) We propose an adaptive weighted local fusion block (LFB) in the nonlinear mapping module, which consists of multiple adaptive weighted residual units (AWRUs) and a local residual fusion unit (LRFU) as shown in Figure 2. The adaptive learning weights in AWRU can help the flow of information and gradients more efficiently and effectively, while LRFU can effectively fuse multi-level residual information in LFB.

(2) We propose an AWMS reconstruction module, which can make full use of the features from nonlinear mapping module to improve the reconstruction quality. Moreover, the redundant scale branches can be removed according to the weighted contribution to further reduce network parameters.

(3) The proposed lightweight AWSRN achieves superior SR reconstruction performance compared to state-of-the-art CNN-based SISR algorithms. As shown in Figure 1, the proposed AWSRN algorithm achieves state-of-the-art performance on Set5 with a scale factor of 2. Moreover, the AWSRN family has a good trade-off between reconstruction performance and the number of operations.

## 2. Related Work

### 2.1. Lightweight Super-Resolution Networks

VDSR is a classic SR network after SRCNN [4], which has very deep network depth. Thereafter, the CNN-based SR models show a trend that the deeper the network, the better the performance. For example, MDSR [6] network stacked more than 160 layer networks with the improved residual unit; RCAN [8] built an SR network with more than 400 layers and achieved better SR performance by the residual in residual and channel attention mechanisms. However, the number of parameters and the number of operations are also greatly increased in deeper networks, which limits their real-world applications [9]. Therefore, it is still a challenging task to build the lightweight SR networks

Dong et al. further proposed a faster SRCNN (FSRCNN) algorithm after SRCNN to reduce the computational cost by using the original LR image instead of the interpolated LR image as the input of the SR network [3]. DRRN [11] shared parameters through recursive mechanism to not only reduce the parameters, but also improve the reconstruction quality of SR images. Ahn et al. proposed a cascading residual network (CARN) to achieve lightweight and efficient reconstruction [9]. More recently, the NAS-based SR networks, such as MoreMNA-S [13] and FALSR [14], have shown their efficient performance. All these works suggest that the lightweight SR networks can keep a good trade-off between reconstruction quality and the number of network parameters.

### 2.2. Efficient Residual Learning

RL has been widely used in CNN and its variants for various CV tasks. VDSR is the first RL-based SR network that significantly improves the reconstruction quality compared with SRCNN[4]. Thereafter, SRResNet adopted the residual unit in ResNet to build a deeper network [17]. In NTIRE 2017 Super Resolution Challenge [18], EDSR [6] algorithm achieved the best performance by further improving the residual unit of SRResNet. Moreover, WDSR won the NTIRE 2018 Super Resolution Challenge with the wide-activate residual unit [19,20]. All these works indicate the effectiveness of RL.

On the other hand, Jung et al. proposed a weighted residual unit (wRU) to improve the performance of image classification[21], which designed a similar network module to Squeeze-and-Excitation (named wSE hereafter) [22], so as to generate weights for residual unit. As shown in Figure 3(b), this wSE strategy can be regarded as an extended form of residual scaling. Therefore, it is feasible to apply wSE to SR network for superior reconstruction. However, wSE results in additional parameters and computational overhead to generate weights.

Motivated by wSE, we propose a novel adaptive weighted residual unit, namely AWRU, to adaptively learn weight for the residual unit in SR network. As shown in Figure 3(c), our AWRU is different from the wSE, where the weights are independently generated without any additional module in network. Therefore, there is no additional computational overhead introduced by AWRU, which makes it easy to be applied to various residual structures.

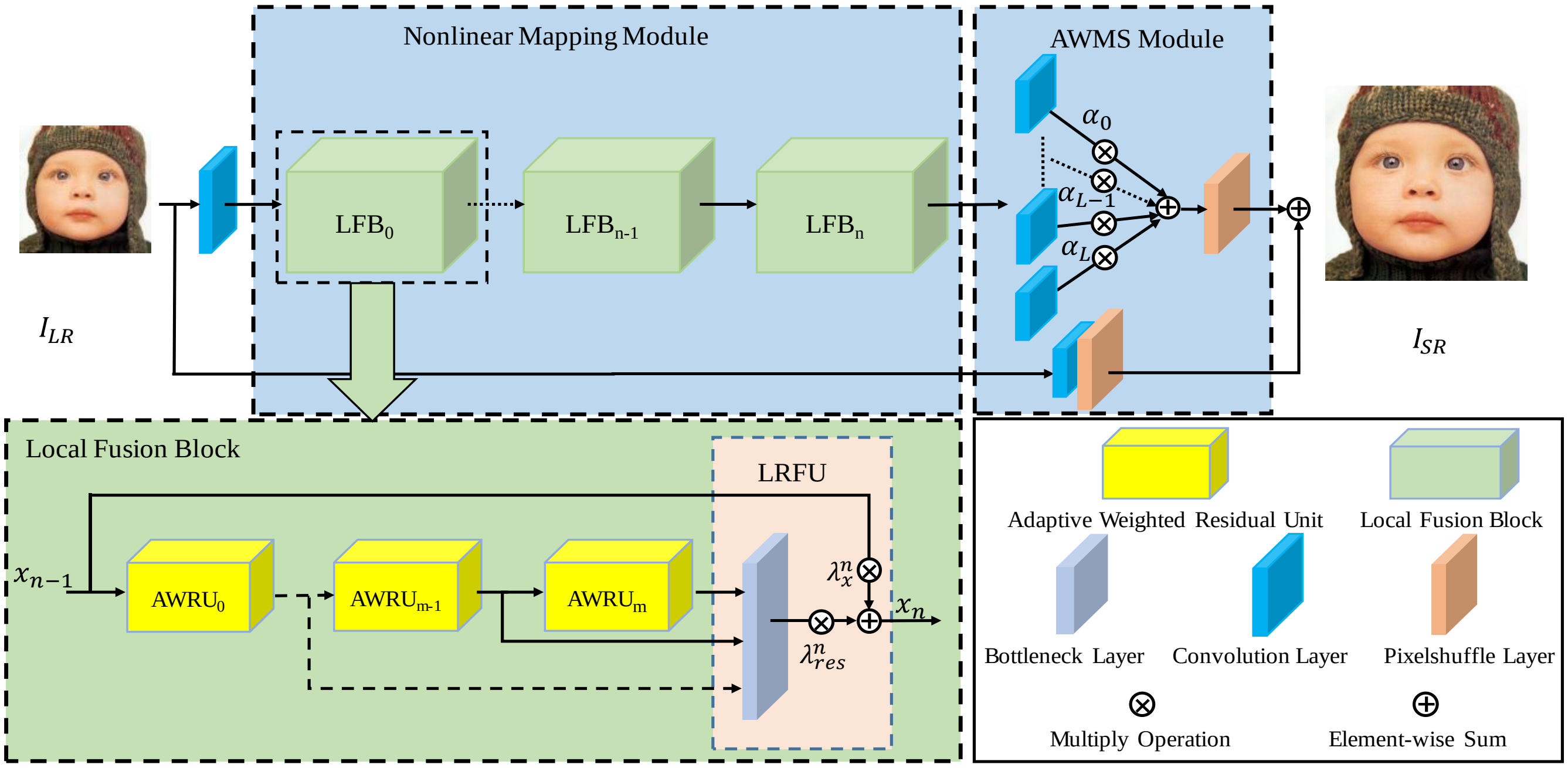

Figure 2. Network architecture of the proposed AWSRN

### 2.3. Upsampling Layer

In SISR, upsampling is also an important factor that affect the SR reconstruction. Interpolation is a commonly used method in SR networks, such as SRCNN, VDSR and DRRN, to resize the original LR image to the target size as the input of a CNN model for SR reconstruction [1,4,11]. However, the computational operations are greatly increased due to the large size of the input image. Therefore, FSRCNN and SRDenseNet directly adopted the original LR image without upsampling as input for CNN, in which a transposed convolution layer was added to implement the final upsampling reconstruction [23]. This method greatly reduces unnecessary computational overhead. Furthermore, EPSCN proposed a method called subpixelshuffle to overcome the problem of the checkerboard effect in transposed convolution. Subpixelshuffle has been widely used in recently proposed SR models, such as EDSR [6], WDSR [19] and RCAN [8]. These models only use a single-scale module for reconstruction, which does not fully utilize the feature information from nonlinear mapping layer. On the other hand, although the multi-scale reconstruction can achieve superior reconstruction quality, it generally results in more parameters and computational overhead.

In this work, we propose an AWMS reconstruction module with multiple scale convolutions to the trade-off between the reconstruction quality and module parameters. That is, we can remove some scale branch with lower contribution according to the automatically learned weights to reduce the parameters while without losing performance

## 3. Method

### 3.1. Basic Network Architecture

As shown in Figure 2, the proposed AWSRN consists of three modules, namely the feature extraction module, the nonlinear mapping module stacked with several LFBs, and the AWMS reconstruction module.

The feature extraction module is a convolution layer with a kernel size of 3×3, which can be formulated as

$$x_0 = f_{ext}(I_{LR}) \tag{1}$$

where $f_{ext}$ denotes the feature extraction function for the LR image $I_{LR}$ and $x_0$ is the output feature map from the first convolutional layer.

Define the proposed LFB as $f_{LFB}$. The non-linear mapping module is stacked with several LFBs, given by

$$x_n = f_{LFB}^{n}(f_{LFB}^{n-1}(\dots f_{LFB}^{0}(x_0)\dots)) \tag{2}$$

where $x_n$ is the output feature of the $n$th LFB.

The output $x_n$ of the last LFB is then fed to the AWMS reconstruction module. In addition, we implement the global residual path $f_{up}$ in the AWMS module by stacking a convolution layer and a subpixelshuffle layer, which uses $I_{LR}$ as input.

$$I_{SR} = \sum_{i=1}^{M} \alpha_i f_{rec}^{i}(x_n) + f_{up}(I_{LR}) \tag{3}$$

where $f_{rec}$ is the muti-scale reconstruct module, $\alpha_i$ is the adaptive weight of the $i$th scale branch of the multi-scale reconstruction module.

### 3.2. Local Fusion Block

The nonlinear mapping module is stacked by several LFBs, while a LFB consists of two parts: multiple stacked AWRUs and one LRFU.

Define $x_{m-1}$ and $x_m$ as the input and output of the $m$th LFB, respectively, and $\lambda_{res}^m$ and $\lambda_x^m$ are the corresponding adaptive weights, respectively. LFB can be expressed as

$$x_m = \lambda_{res}^m f_{red}^m([x_m^0, x_m^1, \ldots, x_m^n]) + \lambda_x^m x_{m-1} \quad (4)$$

where $x_m^n$ is the output of the $n$th AWRU in the LFB, and $f_{red}^m$ represents the bottleneck function of the $m$th LFB.

We employ the wide-activate residual unit in WDSR shown in Figure 3(a) as our basic residual unit (Basic RU). This unit allows more low-level information to be activated without increasing parameters by shrinking the dimensions of the input/output and extending the internal dimensions before ReLU [19].

We then propose the AWRU based on the Basic RU. As shown in Figure 3(c), AWRU contains only two independent weights, which can be adaptively learned after they are given an initial value. On the contrary, the wRU in Figure 3(b) generates two weights for the residual unit through wSE, which results in more parameters.

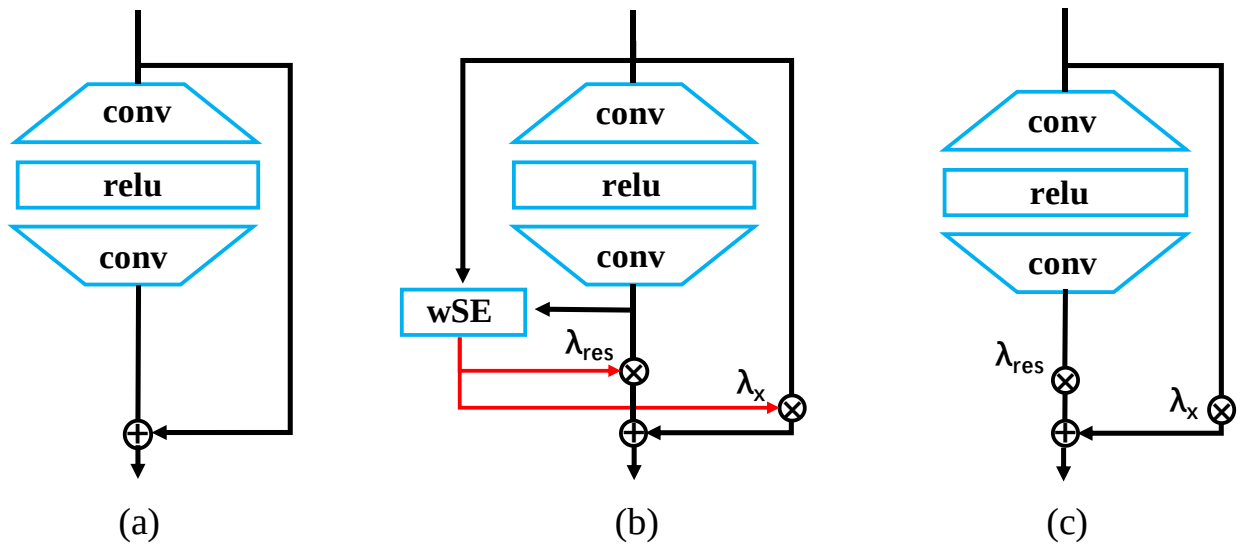

Figure 3. (a) Basic RU from WDSR that does not have any weight; (b) wRU that generates two weights for the residual unit with wSE; (c) Our proposed AWRU that has two independent weights for the residual unit.

Define $x_{k-1}$ and $x_k$ as the input and output feature map of residual unit, respectively. The Basic RU can be written as

$$x_k = f_{res}^k(x_{k-1}) + x_{k-1} \quad (5)$$

where $f_{res}^k$ is the $k$th residual function. Both wRU and our AWRU can be written as

$$x_k = \lambda_{res}^k f_{res}^k(x_{k-1}) + \lambda_x^k x_{k-1} \quad (6)$$

where $\lambda_{res}^k$ and $\lambda_x^k$ are the weight values for two branches of the residual unit.

In order to make better use of the feature information of AWRUs in LFB, the LRFU is then proposed to fuse multi-level feature information. As shown in Figure 2, the bottleneck layer is added in LRFU to fuse multiple levels of information and also to match the dimensions of shortcut branches.

## 4. Experiments

### 4.1. Datasets and Metrics

The most popular used dataset DIV2K was selected to train the proposed AWSRN in this work. This dataset includes 800 pairs of pictures where the LR image is obtained by the bicubic downsampling of HR image [20]. During testing stage, several standard benchmark datasets, namely Set5 [24], Set14 [25], B100 [26], Urban100 [28], Manga109 [27], were used for evaluation. The peak signal to noise ratio (PSNR) and the structural similarity index (SSIM) [29] on the Y channel after converting to YCbCr channels were calculated as the evaluation metrics.

### 4.2. Implementation Details

We randomly cropped 16 patches of size 48×48 from the LR images as input for each training minibatch. Data augmentation was performed on the training set, such as random rotations of 90°, 180°, 270° and horizontal flips.

For the setting of hyperparameters, we set {32,128,32} channels for AWRU, which means the input, internal, output channel number is 32,128,32, respectively. The initial values of the adaptive weights in all AWRUs are 1. The LFB had 4 AWRUs by default. For AWMS module, we set 4 scale layers with the kernel sizes of 3×3, 5×5, 7×7, and 9×9, respectively. The initial weight of each scale branch is 0.25. Our models were trained by ADAM optimizer [30] with L1 loss [31]. The learning rate was set to $10^{-3}$ by using weight normalization and then decreased to half every $2\times 10^5$ iterations of back-propagation. We implemented our model using PyTorch framework [32] with an NVIDIA 1080Ti GPU.

### 4.3. Ablation Study

**Weighted Residual Unit.** To evaluate the effectiveness of the weighted residual unit on SR reconstruction, we used WDSR with 16 residual units as the basic network, and then replaced the residual units in WDSR by the Basic RU, wRU and AWRU, respectively. We implemented those models on all benchmark datasets with the scale factor of 2, and calculated the mean PSNR value to evaluate the effect of WDSR.

Table 1 gives the compared results of different RUs in WDSR. It can be found that both wRU and AWRU achieve superior performance with higher PSNR values compared to the Basic RU, suggesting the effectiveness of the weighted residual unit. Moreover, our AWRU has fewer parameters than wRU.

| RU | Basic RU | wRU | AWRU |
|---|---|---|---|
| Params diff. | 0 | +8768 | +32 |
| PSNR | 34.99 | 35.02 | 35.02 |

Table 1. The effect of WDSR with different residual units on the difference in parameter number and PSNR.

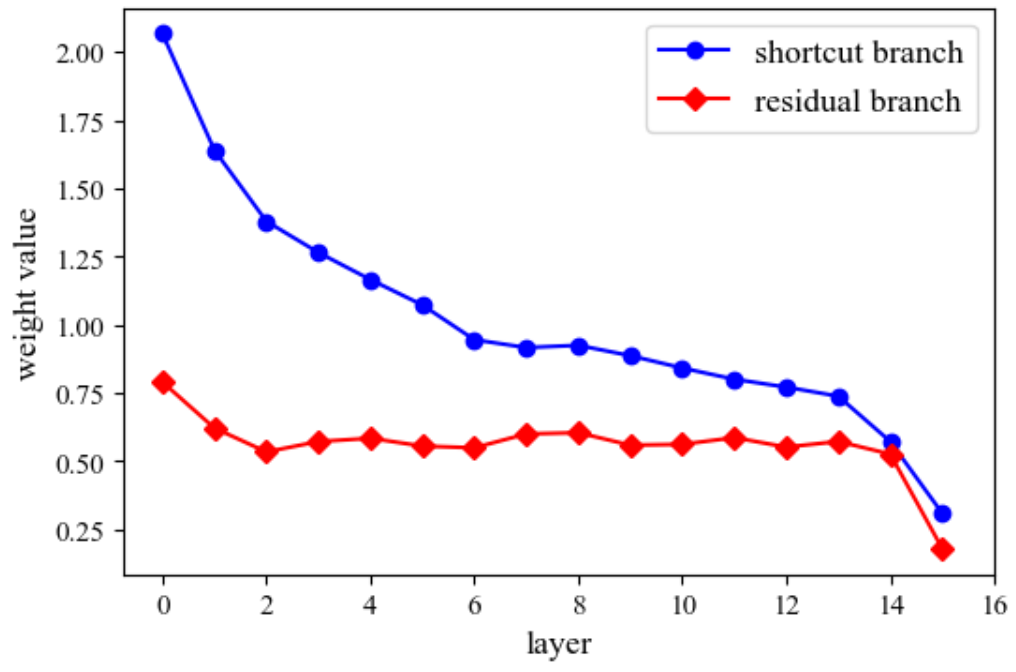

Figure 4. The weight of the residual branch and the shortcut branch of the ARWU in different layers of WDSR.

Figure 4 shows the weight $\lambda_{res}$ of the AWRU residual branch and the weight $\lambda_x$ of the shortcut branch in the AWRU-based WDSR model. It can be observed that the weight values of both the residual branch and the shortcut branch decrease with the increasing of the network depth. It suggests that the deeper AWRU in the network requires a smaller scale value to prevent from gradient explosion. Therefore, the role of residual scaling is more important in the deeper layer of network. Besides, the AWRU has greater weights in the shallow layers, especially in the shortcut branch, which is totally different from the commonly used trick of residual scaling. It means that more information in the shallow layer network needs to be transmitted to the deeper layers of the network.

| | baseline | AWSRN-NA | WASRN-NL | AWSRN-B |
|---|---|---|---|---|
| AWRU | × | × | ✓ | ✓ |
| LRFU | × | ✓ | × | ✓ |
| PSNR | 34.99 | 35.03 | 35.02 | 35.08 |

Table 2. Results of ablation study on effects of AWRU and LRFU.

**The Importance of AWRU and LRFU.** To evaluate the performance of the AWRU and LRFU components in LFB, we first set a model with 16 Basic RU as the baseline model and an AWSRN-B with 4 LFBs stacked. AWSRN-B has the same number of RUs as the baseline. Then, to observe the performance of the AWRU for reconstruction, we set up the model AWSRN-NA, which replaced the AWRU in each LFB with a Basic RU. To observe the performance of the LRFU, we set up the model AWSRN-NL, which removed the LRFU from each LFB. PSNR is the calculated average of all benchmark datasets with the scale factor of 2. As shown in Table 2, AWSRN-B outperforms both AWSRN-NA and AWSRN-NL, indicating the effectiveness of our proposed AWRU and LRFU.

**Adaptive Weighting Multi-Scale Reconstruction.** To demonstrate the effectiveness of AWMS module in the reconstruction layer, we analyzed the performance on WDSR with 16 Basic RU and AWSRN-B with 4 LFBs. As shown in Table 3, our AWMS module showed superior performance to the reconstruction layer that only had a 3×3 convolutional kernel. We implemented those models on all benchmark datasets with the scale factor of 2, and calculated the mean PSNR

| backbone | WDSR | | AWSRN-B | |
|---|---|---|---|---|
| AWMS | × | ✓ | × | ✓ |
| PSNR | 34.99 | 35.02 | 35.08 | 35.1 |

Table 3. Results of ablation study on effects of AWMS module with different backbones.

Furthermore, it can be found that the weights on different scale branches have different contributions. This means that different feature information can be obtained in our AWMS module.

| backbone | | | WDSR w/ AWMS | | | | |
|---|---|---|---|---|---|---|---|
| weight of kernel | 3 | 0.1029 | × | | | | ✓ |
| | 5 | 0.0190 | | × | | | ✓ |
| | 7 | 0.0111 | | | × | | ✓ |
| | 9 | 0.0088 | | | | × | ✓ |
| PSNR | | | 32.92 | 38.07 | 38.09 | 38.09 | 38.09 |
| backbone | | | AWSRN-B w/ AWMS | | | | |
| weight of kernel | 3 | 0.1282 | × | | | | ✓ |
| | 5 | 0.0211 | | × | | | ✓ |
| | 7 | -0.0003 | | | × | | ✓ |
| | 9 | 0.0173 | | | | × | ✓ |
| PSNR | | | 32.37 | 38.09 | 38.11 | 38.10 | 38.11 |

Table 4. Results of ablation study on effects of removing different scale branches in reconstruction layer with different backbones. PSNR is the calculated average of Set5 with the scale factor of 2.

In order to further analyze the effects of different kernel sizes on reconstruction quality, we removed the features on each branch respectively and tested the results on Set5. As shown in Table 4, It suggests that the branches with 3×3 and 5×5 kernel sizes have more impacts on the results. The branches with 7×7 and 9×9 kernel on WDSR have smaller weights. After removing branch of 7×7 kernel, there is no significant effect on the reconstruction quality neither for WDSR nor AWSRN-B. Figure 5 shows the visualization results on different scale branches. It can be found that the 3×3 scale branch mainly captures low-frequency information. The large-scale branches in AWMS module are more sensitive to the high-frequency information. It can be seen that our AWMS module has a good ability to capture both low frequency and high frequency information for reconstruction.

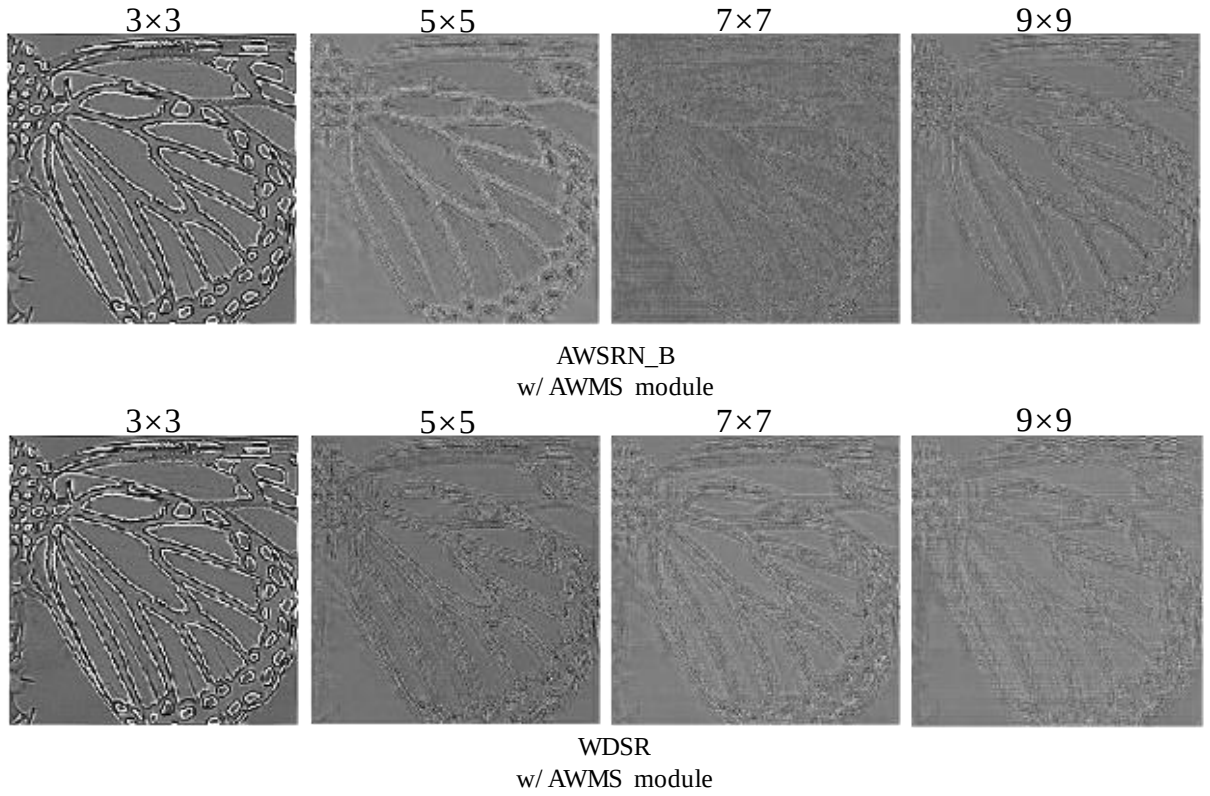

Figure 5. Visualization results on different scale branches in AWMS, 3×3 scale branch has more low frequency information, and other scales show more high frequency edge information.

### 4.4. Comparison with State-of-the-art Methods

We compared the proposed AWSRN with many lightweight and efficient SR methods on ×2, ×3, ×4, ×8 scales, including SRCNN [1], FSRCNN [3], VDSR [4], DRCN [5], LapSRN [34], DRRN [11], SelNet [35], MemNet [36], CARN [9], MoreMNAS-A [13], FALSR [14], SRMDNF [37], MSRN [38].

For a comprehensive comparison, we designed four models, named AWSRN-S, AWSRN-SD, AWSRN-M and AWSRN. That is, for AWSRN-S, AWSRN-M and AWSRN, we stacked 1, 3, and 4 LFBs, respectively. Each LFB has 4 AWRUs, and AWRU has {32,128,32} channels.

Since AWSRN-S has only 8 layers in one LFB, we set a deeper version for AWSRN-S, named AWSRN-SD, which also has only one LFB but has 8 AWRUs, each AWRU has {16,128,16} channels.

Table 5 shows the results of PSNR and SSIM on five benchmark datasets for different algorithms. In addition, the parameters and Muti-Adds of models are also given for a more intuitive comparison. Muti-Adds was assumed to be calculated with a 1280×720 SR image at all scales. In particular, it is worth noting that both MoreMNAS-A and FALSR only have results on ×2 scale, and CARN does not have result on ×8 scale. The results show that our models have achieved the best performance in each parameter scale.

Our AWSRN-S, which has the same small parameters and Muti-Adds as CARN-M [9], FLASR-B [14], and FLASR-C [14], achieves better results than these small models on ×2 and ×3. Only on ×4, CARN-M is slightly better than AWSRN-S due to its deeper network. However, our AWSRN-SD, which is a deeper version of AWSRN-S and has fewer parameters and Muti-Adds than AWSRN-S, achieves the best results among all these small models.

On models with more parameters, our AWSRN-M and AWSRN also achieve the best results. Specifically, AWSRN-M shows better performance than SelfNet, MoreMNAS-A [13], FALSR-A [14] with similar parameter numbers and Muti-Adds. And our AWSRN achieves far better performance than CARN with a slightly smaller number of parameters. Although the Muti-Adds of AWSRN is slightly higher than CARN, better results are achieved compared to MSRN that has higher Muti-Adds and more than 6000K parameters. In addition, it should be noted that we do not remove redundant scale branch in AMWS module for the results, which means that we can further reduce the number of parameters and the number of operations in our model. In Figure 6, we illustrated the visual comparisons over four datasets (Set5, Set14, B100 and Urban100) for × 4 scale. We selected some detail patches from images. It can be seen that the SR image reconstructed by our model is closer to the HR image in details.

It is worth noting that EDSR [6], D-DBPN [33], RDN [7] and RCAN [8] have higher performance than our AWSRN family, but these models have more than 10M parameters. we have the highest number of parameters on the x8 scale of AWSRN while the number of parameters is only 2348K, which is less than 5% of the number of EDSR parameters.

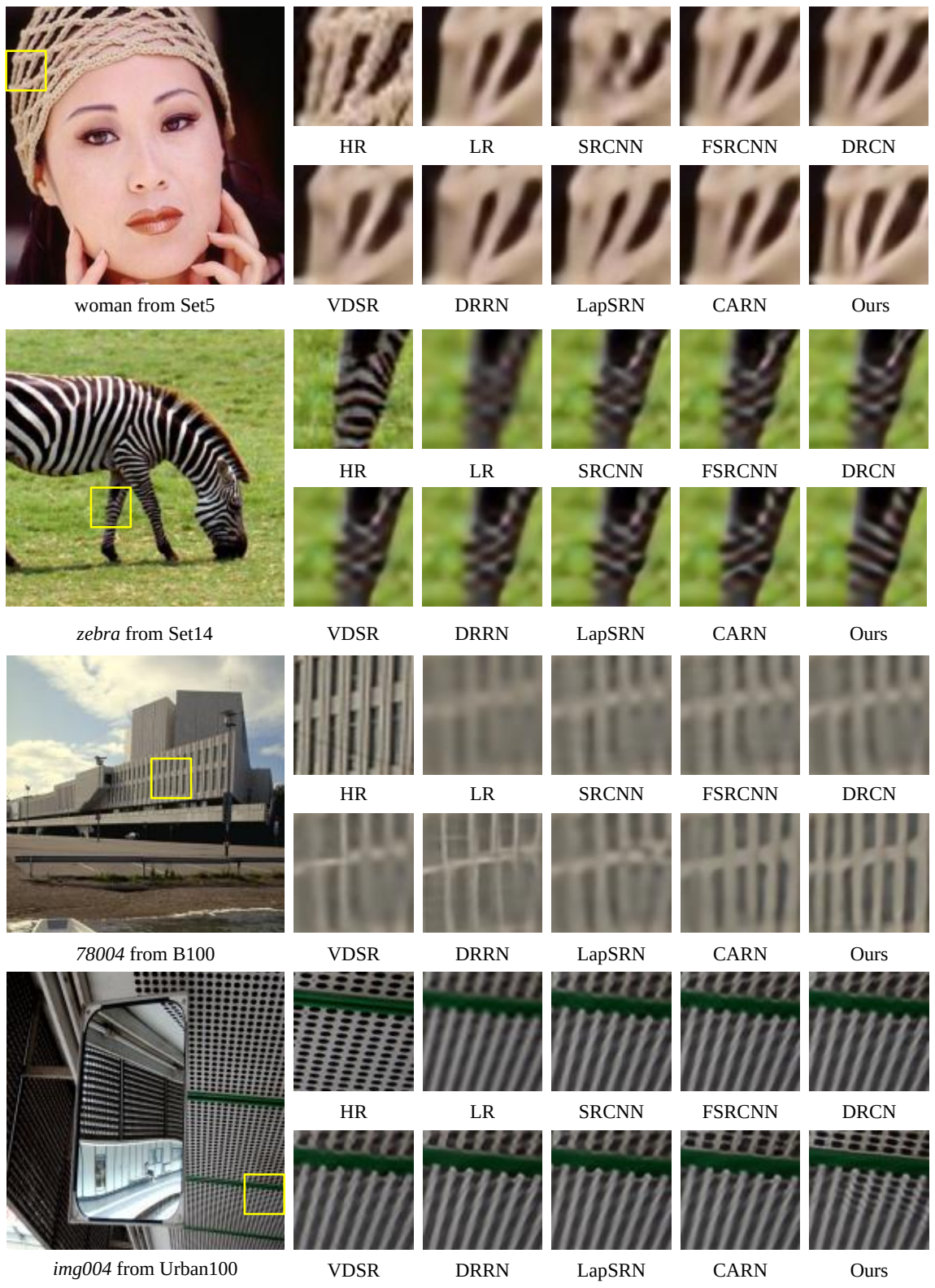

Figure 6. Visual comparisons over four datasets (Set5, Set14, B100 and Urban100) for ×4 scale.

| Scale | Model | Params | MutiAdds | Set5 PSNR/SSIM | Set14 PSNR/SSIM | B100 PSNR/SSIM | Urban100 PSNR/SSIM | Manga109 PSNR/SSIM |
|---|---|---|---|---|---|---|---|---|
| 2 | SRCNN[1] | 57K | 52.7G | 36.66/0.9542 | 32.42/0.9063 | 31.36/0.8879 | 29.50/0.8946 | 35.74/0.9661 |
| | FSRCNN[3] | 12K | 6.0G | 37.00/0.9558 | 32.63/0.9088 | 31.53/0.8920 | 29.88/0.9020 | 36.67/0.9694 |
| | VDSR[4] | 665K | 612.6G | 37.53/0.9587 | 33.03/0.9124 | 31.90/0.8960 | 30.76/0.9140 | 37.22/0.9729 |
| | DRCN[5] | 1,774K | 17,974G | 37.63/0.9588 | 33.04/0.9118 | 31.85/0.8942 | 30.75/0.9133 | 37.63/0.9723 |
| | LapSRN[34] | 813K | 29.9G | 37.52/0.9590 | 33.08/0.9130 | 31.80/0.8950 | 30.41/0.9100 | 37.27/0.9740 |
| | DRRN[11] | 297K | 6,796.9G | 37.74/0.9591 | 33.23/0.9136 | 32.05/0.8973 | 31.23/0.9188 | 37.92/0.9760 |
| | MemNet[36] | 677K | 2,662.4G | 37.78/0.9597 | 33.28/0.9142 | 32.08/0.8978 | 31.31/0.9195 | - |
| | CARN-M[9] | 412K | 91.2G | 37.53/0.9583 | 33.26/0.9141 | 31.92/0.8960 | 31.23/0.9193 | - |
| | FALSR-B[14] | 326k | 74.7G | 37.61/0.9585 | 33.29/0.9143 | 31.97/0.8967 | 31.28/0.9191 | - |
| | FALSR-C[14] | 408k | 93.7G | 37.66/0.9586 | 33.26/0.9140 | 31.96/0.8965 | 31.24/0.9187 | - |
| | **AWSRN-S (Ours)** | **397K** | **91.2G** | **37.75/0.9596** | **33.31/0.9151** | **32.00/0.8974** | **31.39/0.9207** | **37.90/0.9755** |
| | **AWSRN-SD (Ours)** | **348K** | **79.6G** | **37.86/0.9600** | **33.41/0.9161** | **32.07/0.8984** | **31.67/0.9237** | **38.20/0.9762** |
| | SelNet[35] | 974K | 225.7G | 37.89/0.9598 | 33.61/0.9160 | 32.08/0.8984 | - | - |
| | MoreMNAS-A[13] | 1,039K | 238.6G | 37.63/0.9584 | 33.23/0.9138 | 31.95/0.8961 | 31.24/0.9187 | - |
| | FALSR-A[14] | 1,021K | 234.7G | 37.82/0.9595 | 33.55/0.9168 | 32.12/0.8987 | 31.93/0.9256 | - |
| | **AWSRN-M (Ours)** | **1,063K** | **244.1G** | **38.04/0.9605** | **33.66/0.9181** | **32.21/0.9000** | **32.23/0.9294** | **38.66/0.9772** |
| | SRMDNF[37] | 1,513K | 347.7G | 37.79/0.9600 | 33.32/0.9150 | 32.05/0.8980 | 31.33/0.9200 | - |
| | CARN[9] | 1,592K | 222.8G | 37.76/0.9590 | 33.52/0.9166 | 32.09/0.8978 | 31.92/0.9256 | - |
| | MSRN[38] | 5,930K | 1365.4G | 38.08/0.9607 | 33.70/0.9186 | 32.23/0.9002 | 32.29/0.9303 | 38.69/0.9772 |
| | **AWSRN (Ours)** | **1,397K** | **320.5G** | **38.11/0.9608** | **33.78/0.9189** | **32.26/0.9006** | **32.49/0.9316** | **38.87/0.9776** |
| 3 | SRCNN[1] | 57K | 52.7G | 32.75/0.9090 | 29.28/0.8209 | 28.41/0.7863 | 26.24/0.7989 | 30.59/0.9107 |
| | FSRCNN[3] | 12K | 5.0G | 33.16/0.9140 | 29.43/0.8242 | 28.53/0.7910 | 26.43/0.8080 | 30.98/0.9212 |
| | VDSR[4] | 665K | 612.6G | 33.66/0.9213 | 29.77/0.8314 | 28.82/0.7976 | 27.14/0.8279 | 32.01/0.9310 |
| | DRCN[5] | 1,774K | 17,974G | 33.82/0.9226 | 29.76/0.8311 | 28.80/0.7963 | 27.15/0.8276 | 32.31/0.9328 |
| | DRRN[11] | 297K | 6,796.9G | 34.03/0.9244 | 29.96/0.8349 | 28.95/0.8004 | 27.53/0.8378 | 32.74/0.9390 |
| | MemNet[36] | 677K | 2,662.4G | 34.09/0.9248 | 30.00/0.8350 | 28.96/0.8001 | 27.56/0.8376 | - |
| | CARN-M[9] | 412K | 46.1G | 33.99/0.9236 | 30.08/0.8367 | 28.91/0.8000 | 27.55/0.8385 | - |
| | **AWSRN-S (Ours)** | **477K** | **48.6G** | **34.02/0.9240** | **30.09/0.8376** | **28.92/0.8009** | **27.57/0.8391** | **32.82/0.9393** |
| | **AWSRN-SD (Ours)** | **388K** | **39.5G** | **34.18/0.9273** | **32.21/0.8398** | **28.99/0.8027** | **27.80/0.8444** | **33.13/0.9416** |
| | SelNet[35] | 1,159K | 120.0G | 34.27/0.9257 | 30.30/0.8399 | 28.97/0.8025 | - | - |
| | **AWSRN-M (Ours)** | **1,143K** | **116.6G** | **34.42/0.9275** | **30.32/0.8419** | **29.13/0.8059** | **28.26/0.8545** | **33.64/0.9450** |
| | SRMDNF[37] | 1,530K | 156.3G | 34.12/0.9250 | 30.04/0.8370 | 28.97/0.8030 | 27.57/0.8400 | - |
| | CARN[9] | 1,592K | 118.8G | 34.29/0.9255 | 30.29/0.8407 | 29.06/0.8034 | 28.06/0.8493 | - |
| | MSRN[38] | 6,114K | 625.7G | 34.46/0.9278 | 30.41/0.8437 | 29.15/0.8064 | 28.33/0.8561 | 33.67/0.9456 |
| | **AWSRN (Ours)** | **1,476K** | **150.6G** | **34.52/0.9281** | **30.38/0.8426** | **29.16/0.8069** | **28.42/0.8580** | **33.85/0.9463** |
| 4 | SRCNN[1] | 57K | 52.7G | 30.48/0.8628 | 27.49/0.7503 | 26.90/0.7101 | 24.52/0.7221 | 27.66/0.8505 |
| | FSRCNN[3] | 12K | 4.6G | 30.71/0.8657 | 27.59/0.7535 | 26.98/0.7150 | 24.62/0.7280 | 27.90/0.8517 |
| | VDSR[4] | 665K | 612.6G | 31.35/0.8838 | 28.01/0.7674 | 27.29/0.7251 | 25.18/0.7524 | 28.83/0.8809 |
| | DRCN[5] | 1,774K | 17,974G | 31.53/0.8854 | 28.02/0.7670 | 27.23/0.7233 | 25.14/0.7510 | 28.98/0.8816 |
| | LapSRN[34] | 813K | 149.4G | 31.54/0.8850 | 28.19/0.7720 | 27.32/0.7280 | 25.21/0.7560 | 29.09/0.8845 |
| | DRRN[11] | 297K | 6,796.9G | 31.68/0.8888 | 28.21/0.7720 | 27.38/0.7284 | 25.44/0.7638 | 29.46/0.8960 |
| | MemNet[36] | 677K | 2,662.4G | 31.74/0.8893 | 28.26/0.7723 | 27.40/0.7281 | 25.50/0.7630 | - |
| | CARN-M[9] | 412K | 32.5G | 31.92/0.8903 | 28.42/0.7762 | 27.44/0.7304 | 25.62/0.7694 | - |
| | **AWSRN-S (Ours)** | **588K** | **33.7G** | **31.77/0.8893** | **28.35/0.7761** | **27.41/0.7304** | **25.56/0.7678** | **29.74/0.8982** |
| | **AWSRN-SD (Ours)** | **444K** | **25.4G** | **31.98/0.8921** | **28.46/0.7786** | **27.48/0.7368** | **25.74/0.7746** | **30.09/0.9024** |
| | SelNet[35] | 1,417K | 83.1G | 32.00/0.8931 | 28.49/0.7783 | 27.44/0.7325 | - | - |
| | **AWSRN-M (Ours)** | **1,254K** | **72.0G** | **32.21/0.8954** | **28.65/0.7832** | **27.60/0.7368** | **26.15/0.7884** | **30.56/0.9093** |
| | SRDenseNet[23] | 2,015K | 389.9G | 32.02/0.8934 | 28.50/0.7782 | 27.53/0.7337 | 26.05/0.7819 | - |
| | SRMDNF[37] | 1,555K | 89.3G | 31.96/0.8930 | 28.35/0.7770 | 27.49/0.7340 | 25.68/0.7730 | - |
| | CARN[9] | 1,592K | 90.9G | 32.13/0.8937 | 28.60/0.7806 | 27.58/0.7349 | 26.07/0.7837 | - |
| | MSRN[38] | 6,078K | 349.8G | 32.26/0.8960 | 28.63/0.7836 | 27.61/0.7380 | 26.22/0.7911 | 30.57/0.9103 |
| | **AWSRN (Ours)** | **1,587K** | **91.1G** | **32.27/0.8960** | **28.69/0.7843** | **27.64/0.7385** | **26.29/0.7930** | **30.72/0.9109** |
| 8 | SRCNN[1] | 57K | 52.7G | 25.34/0.6471 | 23.86/0.5443 | 24.14/0.5043 | 21.29/0.5133 | 22.46/0.6606 |
| | FSRCNN[3] | 12K | 4.6G | 25.42/0.6440 | 23.94/0.5482 | 24.21/0.5112 | 21.32/0.5090 | 22.39/0.6357 |
| | VDSR[4] | 665K | 612.6G | 25.73/0.6743 | 23.20/0.5110 | 24.34/0.5169 | 21.48/0.5289 | 22.73/0.6688 |
| | DRCN[5] | 1,774K | 17,974G | 25.93/0.6743 | 24.25/0.5510 | 24.49/0.5168 | 21.71/0.5289 | 23.20/0.6686 |
| | LapSRN[34] | 813K | - | 26.15/0.7028 | 24.45/0.5792 | 24.54/0.5293 | 21.81/0.5555 | 23.39/0.7068 |
| | MSRN[38] | 6,226K | 89.6G | 26.59/0.7254 | 24.88/0.5961 | 24.70/0.5410 | 22.37/0.5977 | 24.28/0.7517 |
| | **AWSRN (Ours)** | **2,348K** | **33.7G** | **26.97/0.7747** | **24.99/0.6414** | **24.80/0.5967** | **22.45/0.6174** | **24.60/0.7782** |

Table 5. Qualitative results on benchmark datasets. **Bold** indicates the results of our model, red color indicates the best result, blue color indicates the second best result.

## 5. Conclusions

In summary, we propose a lightweight and efficient adaptive weighted super-resolution network for SISR. Our models achieve better performance than state-of-the-art algorithms without increasing parameters and computational overhead. The efficiency of the proposed algorithm is mainly from the following two reasons: (1) We design the local fusion block (LFB) for efficient residual learning, in which the proposed adaptive weighted residual unit and local residual fusion unit can allow efficient flow and integration of information and gradient; (2) The proposed adaptive weighted multi-scale (AWMS) reconstruction module can not only make full use of context information, but also analyze the information redundancy between different scale branches for reducing parameters. More importantly, the adaptive weighted methods we propose in network are very simple and effective, which can be flexibly applied to other SR models or other visual tasks. Moreover, we hope that this work can be applied to real-world applications such as live video.